\documentclass[conference]{IEEEtran}

\usepackage{graphicx}
\usepackage{amsmath,amssymb}
\usepackage{siunitx}
\usepackage{upgreek}
\usepackage[noend]{algpseudocode}
\usepackage{algorithm}
\usepackage{multirow}
\usepackage{url}
\usepackage{balance}
\let\labelindent\relax
\usepackage{enumitem}
\usepackage{xcolor}
\usepackage{soul}
\usepackage{hyperref}

\ifCLASSINFOpdf
\else
\fi

\begin{document}
%
\title{Counterfactual Explanations of Neural Network-Generated Response Curves}



\author{\IEEEauthorblockN{Giorgio Morales and John Sheppard}
\IEEEauthorblockA{Gianforte School of Computing \\
Montana State University\\
Bozeman, MT 59717}
}


\maketitle

\begin{abstract}
Response curves exhibit the magnitude of the response of a sensitive system to a varying stimulus.
However, response of such systems may be sensitive to multiple stimuli (i.e., input features) that are not necessarily independent. 
As a consequence, the shape of response curves generated for a selected input feature (referred to as ``active feature") might depend on the values of the other input features (referred to as ``passive features"). 
In this work we consider the case of systems whose response is approximated using regression neural networks.
We propose to use counterfactual explanations (CFEs) for the identification of the features with the highest relevance on the shape of response curves generated by neural network black boxes.
CFEs are generated by a genetic algorithm-based approach that solves a multi-objective optimization problem.
In particular, given a response curve generated for an active feature, a CFE finds the minimum combination of passive features that need to be modified to alter the shape of the response curve.
We tested our method on a synthetic dataset with 1-D inputs and two crop yield prediction datasets with 2-D inputs.
The relevance ranking of features and feature combinations obtained on the synthetic dataset coincided with the analysis of the equation that was used to generate the problem.
Results obtained on the yield prediction datasets revealed that the impact on fertilizer responsivity of passive features depends on the terrain characteristics of each field.
\footnote{This paper is a preprint (accepted to appear in the International Joint Conference on Neural Networks 2023). IEEE copyright notice. 2023 IEEE. Personal use of this material is permitted. Permission from IEEE must be obtained for all other uses, in any current or future media, including reprinting/republishing this material for advertising or promotional purposes, creating new collective works, for resale or redistribution to servers or lists, or reuse of any copyrighted.}
\end{abstract}


\section{Introduction}

Response curves are tools that allow for the analysis of the responsivity of a sensitive system to a particular stimulus; we refer to that stimulus as the ``active feature". Specifically, a response curve is defined as a curve that exhibits the various values taken by the response of the system to all admissible values of the active feature. 

Typically, analysis of response curves is focused on their shape rather than on their absolute values.
For instance, in pharmacology, the shape of a drug's dose–response curve reflects the strength of the drug~\cite{doseresp,doseresp3}.
Likewise, in agriculture, nitrogen fertilizer-yield response (N-response) curves estimate the crop yield based on different fertilizer inputs. 
The shape of such curves can help determining the economic optimum nitrogen rate (EONR), defined as the nitrogen rate beyond which there is no actual profit for the farmers~\cite{bullock94}.

Motivating the current work, response values displayed by response curves may depend, not only on the relationship between the response variable and the active feature, but also on other stimuli, which we will refer to as ``passive features".
Traditionally, such response curves are fitted using univariate (non-)linear regression models between the active feature and the response variable.
Such models could be based on parametric response functions that assume plateau-type, quadratic, and exponential behavior, in the case of N-response curves~\cite{bullock94,watkins},
or sigmoidal, U-shape, and hill functions, in the case of dose-response curves~\cite{doseresp3,doseresp4}.

The traditional approaches assume that, given an active feature, the shape of its response curves is independent of other variables that could possibly affect the absolute response value.
In the case of agricultural applications, fitting a single N-response curve for an entire field implies that the field is homogeneous and behaves similarly at every location.
Nevertheless, recent works suggest that the N-response functional form varies across the fields given the variability of factors such as terrain slope and soil composition~\cite{maxwell,nresponse}. 

In this work, we build upon the conclusion that the shape of response curves might be dependent not only on the active feature but also on the interaction of the active feature with the passive features.
We present a method to derive non-parametric response curves from observed data.
That is, we learn the functional form of the response curves instead of using parametric curve-fitting approaches.
To do this, we train multivariate regression neural networks (NNs) that act as mappings from the feature space and the response value space.
Thus, they are used to generate approximated response curves given an active feature.

The high complexity of the non-linear functions learned by NNs often prevents humans from explaining their behavior, which is why they are usually referred to as ``black-box models."
Given that the importance of NNs in inference tasks has grown rapidly, the area of explainable machine learning (XML) has gained more interest in recent years.
XML aims to allow humans to identify cause-effect relationships between inputs and outputs of black-box models~\cite{XAIreview}.
In concordance with this, we propose a \textit{post-hoc} explainability method that allows for understanding the impact that interacting passive features have on the shape of NN-generated response curves.

\begin{figure}[!t]
    \centering
    \includegraphics[width=7.5cm]{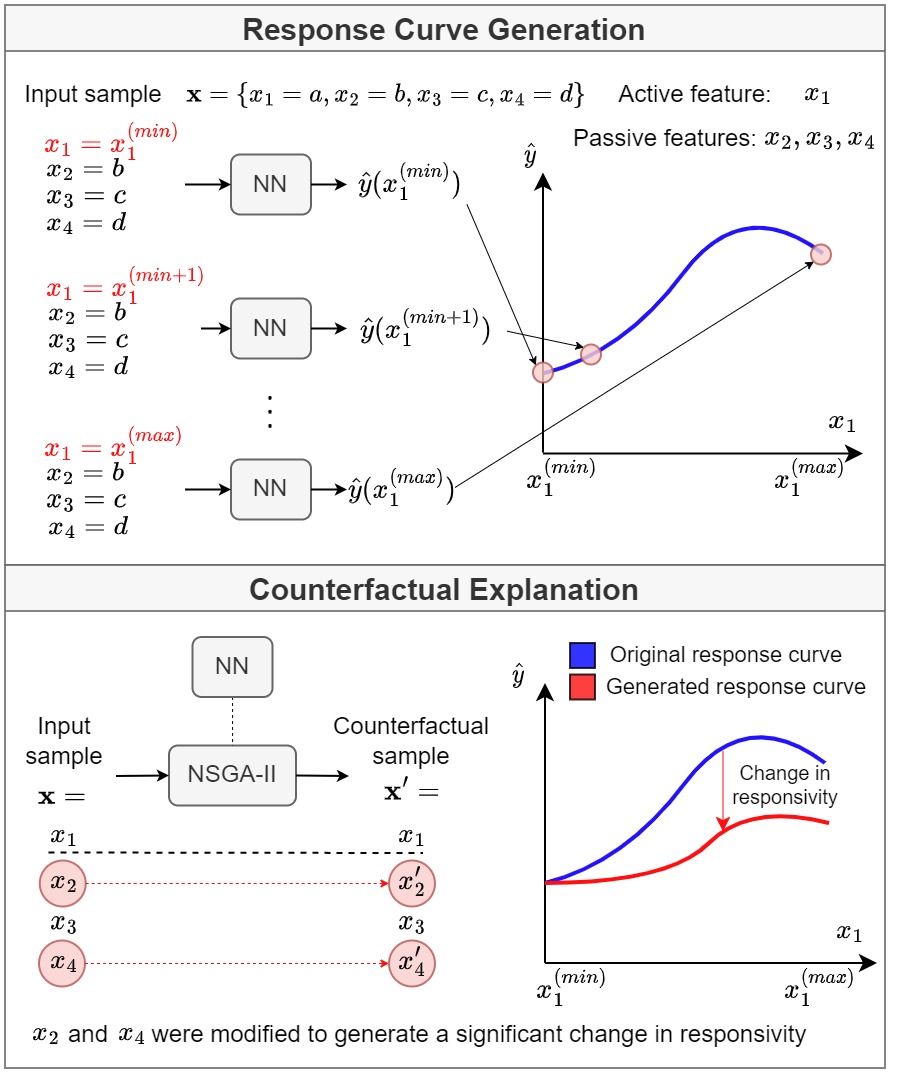}
    \vspace{-1ex}
    \caption{Overview of counterfactual response curve generation methodology.}
    \vspace{-1ex}
    \label{fig:intro}
\end{figure}

Thus, we propose a method that generates counterfactual explanations (CFEs) for each sample to find the minimum number of passive features we need to modify for the response curve of the counterfactual sample to show a change in responsivity with respect to the original one.
The CFE generation problem is posed as a multi-objective optimization (MOO) problem and is solved using a genetic algorithm-based approach.
The upper and lower regions of Fig.~\ref{fig:intro} illustrate the response curve generation and counterfactual response curve generation processes, respectively, of a given sample.
Here, the Non-dominated Sorting Genetic Algorithm II (NSGA-II) represents the selected genetic algorithm.
Finally, we repeat this process for each sample in a dataset to derive global relevance scores, not only for each passive feature but also for specific combinations of them.
Our specific contributions are summarized as follows:
\begin{enumerate}
    \item Our main contribution is a CFE generation method posed as a MOO problem that, given a sample, finds the passive features with the greatest impact on the responsivity of the active feature.
    \item Considering a multivariate regression problem, this is the first work that analyzes the influence of a set of passive features over the shape of the response curves generated for the response variable and a selected active feature.  
    \item We present a method to generate aligned approximate response curves using regression NNs.
    \item We provide global scores that assess the relevance of individual features and combinations of features.
\end{enumerate}

\section{Related Work} \label{sec:related}

Counterfactual explanations are used to identify ``actionable knowledge," i.e., knowledge about causal
dependencies between inputs and outputs of a system~\cite{actionable}.
Such explanations contribute to understanding what could be changed in the input to achieve a desired outcome.
Most CFE methods focus on classification problems where the objective is to find CFEs that produce class labels different from those that were originally predicted~\cite{CFEclass}.
These methods can be categorized as model-agnostic or model-specific.
The former are based on general principles that use outputs of already fitted black-box models~\cite{actionable,CFEMOO,FACE}, where the latter are specific to a particular class of methods, such as differentiable models~\cite{CFEproto} or tree-based ensembles~\cite{CFERF}. 

To date, little attention has been given to counterfactual methods for regression.
Spooner \textit{et al.}~\cite{CFEReg} proved that verifying the existence of counterfactuals is NP-complete.
Regression problems may deal with multiple-dimensional real outputs and, thus, it is not always possible to have efficient algorithms for establishing the existence of counterfactuals with a desired target value.
However, it is not always necessary to find CFEs that lead to specific target values.
For example, Kommiya Mothilal \textit{et al.}~\cite{CFErelev} proposed a method that assesses individual feature relevance from CFEs.
They consider that a feature is more relevant than others for a prediction task if it is changed more often when generating CFEs.

On the other hand, analyses of systems' response curves consider that the shape of the response curve can be described solely by the relationship between the response variable and the active feature~\cite{bullock94,watkins}.
This requires the strong assumption that the explanatory features are mutually independent. 
If this held, the set of passive features would only shift the response curves vertically but would not alter their shape.
However, there is no guarantee for this assumption to always hold, as seen in some precision agriculture applications~\cite{maxwell,nresponse}.
Thus, we argue that it is essential to acknowledge all possible sources of variability in analyzing the shape of response curves. 
To the best of our knowledge, no work has been published on this type of analysis.

In this paper, we explore the use of CFEs to analyze the impact of passive features on the shape of response curves.
It is worth noting that Schwab \textit{et al.}~\cite{CFRresponse} proposed a method to learn counterfactual representations for estimating individual dose-response curves using NNs.
They aimed to estimate what would have happened if a different treatment (dose) had been given to a patient, hence their objective was to produce accurate response estimates across the entire range of all available treatment options.
In this context, the term ``counterfactual outcome" simply refers to the estimated response value obtained for dosages different from the one that the patient actually received (i.e., estimated response curves).
This differs from our perspective significantly.
Unlike previous counterfactual methods that aim to obtain a different response value, our objective is to generate counterfactual samples for a subset of the input features (i.e., passive features) in order to alter the response curve's shape. 
In other words, our CFEs aim to change the system's responsivity; that is, the way it reacts to all admissible values of the active feature.


\section{Methodology}

Let $f(\cdot)$ denote the underlying function of a system whose response $y$ is sensitive to $n$ stimuli (i.e., input features) $\textbf{x} = \{ x_1, \dots, x_n\}$, such that $y = f(\textbf{x}) + \varepsilon$.
Here, $\varepsilon$ is a random variable that represents the error term.
For convenience, we first consider the case where each input and output is one-dimensional ($x_i \in \mathbb{R}, \, \forall i \in [1, n]$ and $y \in \mathbb{R}$).
Later in Sec.~\ref{sec:yieldpred}, we extend this approach to consider two-dimensional inputs.

Without loss of generality, suppose we select the $s$-th feature ($s \in [1, n]$) as the active feature.
The set of remaining features (i.e., passive features) is denoted as $\mathbf{p}=\mathbf{x}\setminus \lbrace x_i\rbrace$.
Thus, a response curve generated for the $s$-th feature, $R_s(\textbf{x})$, consists of the set of values taken by the response $y$ for all admissible values of $x_s$ (bounded by $x^{\min}_s \le x_s \le x^{\max}_s$) as follows:
\[
R_s(\textbf{x}) = \{ f(\textbf{x} | x_s = x^{\min}_s), \dots , f(\textbf{x} | x_s = x^{\max}_s)\}.
\]

\subsection{Response Curve Generation} \label{sec:rcurve}

In many real-life settings, the underlying function $f(\cdot)$ cannot be retrieved directly; therefore, we need to approximate it based on observed data. 
Let $\textbf{X}= \{ \textbf{x}_1, \dots , \textbf{x}_N \}$ be a data set with $N$ training samples, where each sample is denoted as $\textbf{x}_j = \{ x_{j1}, \dots, x_{jn}\}$, and $\textbf{y}= \{ y_1, \dots , y_N \}$ is the set of corresponding target observations.
We construct a NN regression model that captures the association between $\textbf{X}$ and $\textbf{y}$.
Its computed function is denoted as $\hat{f}(\cdot)$, and $\boldsymbol{\theta}$ denotes its weights. 
Thus, given an input $\textbf{x}_j$, the target estimate is computed as $\hat{y}_j = \hat{f}(\textbf{x}_j, \boldsymbol{\theta})$.

The network $\hat{f}$ is trained to reduce the mean square error of the estimations such that the parameters $\boldsymbol{\theta}$ are obtained by the following optimization:
$
\boldsymbol{\theta}^* = \; \underset{\boldsymbol{\theta}}{\text{argmin}} \ \frac{1}{N} \sum_{j=1}^N (\hat{y}_{j} - y_{j})^2.
$
Hence, once the network is trained, and assuming that it captures the underlying causal structure of the problem sufficiently well, it can be used to generate an approximate response curve, $\hat{R}_s(\textbf{x}_j)$, for a given active feature and input sample $\textbf{x}_j$:
\begin{equation}
    \hat{R}_s(\textbf{x}_j) = \{ \hat{f}(\textbf{x}_j | x_{js} = x^{\min}_s), \dots , 
    \hat{f}(\textbf{x}_j | x_{js} = x^{\max}_s)\}.
    \label{eq:response_curve}
\end{equation}

Note that we are not interested in the absolute estimated response values when comparing the shape of two or more curves.
Thus, we get rid of any vertical shifts and obtain the aligned approximate response curve $\tilde{R}_s(\textbf{x}_j)$ by subtracting from $\hat{R}_s(\textbf{x}_j)$ its minimum value:
\begin{equation}
\tilde{R}_s(\textbf{x}_j) = \hat{R}_s(\textbf{x}_j) - \min(\hat{R}_s(\textbf{x}_j)).
\label{eq:align}
\end{equation}
Finally, we create the set $\textbf{R}_s$ that consists of the approximate response curves of all samples in $\textbf{X}$; that is, $\textbf{R}_s = \{ \tilde{R}_s(\textbf{x}_0), \dots , \tilde{R}_s(\textbf{x}_N)\}$.

It is worth mentioning that, previously, we also experimented with other types of classifiers (i.e., support vector machines and random forests) to use in the response curve generation process.
However, due to the fast convergence rates and the substantial improvements in performance, we decided to focus on feedforward NNs (FNNs) and convolutional neural networks (CNNs) for 1-D and 2-D regression, respectively, over the other types of models.  

\subsection{Functional Principal Component Analysis} \label{sec:fPCA}

The set $\textbf{R}_s$ can be interpreted as a set of functional data where each of its samples corresponds to an approximated response curve.
Recall that one of our objectives is to generate counterfactual response curves whose shapes differ with respect to the original ones.
As such, we need to determine a distance metric that conveys the difference in shape between two functional data samples.

In that sense, we use functional principal component analysis (fPCA), which is a tool to extract the dominant modes of variation of functional data~\cite{FDA}.
This approach allows us to obtain a reduced set of $K$ orthonormal functions $\lbrace \xi_1, \ldots \xi_K \rbrace$ so that each curve in $\textbf{R}_s$ is approximated as an expansion of these basis functions as
$
\tilde{R}_s(\textbf{x}_j) \approx \sum_{k=1}^K v_k^{(s)}(\textbf{x}_j) \, \xi_k,
$
where $v_k^{(s)}(\textbf{x}_j)$ is the value of the $k$-th principal component corresponding to the response curve generated for $\textbf{x}_j$ considering the $s$-th feature as the active feature.

A \textit{functional} principal component (fPC) represents a distinct curve pattern.
Hence, two curves with different shapes will be encoded using different fPC values.
We consider using $K=3$ fPCs, as they were sufficient to explain at least 99.5\% of the variance of the datasets used in our experiments. 
We define our distance metric $d_s(\textbf{x}_j, \textbf{x}_q)$, calculated between the transformed response curves generated for $\textbf{x}_j$ and $\textbf{x}_q$, as:
\begin{equation}
    d_s(\textbf{x}_j, \textbf{x}_q) = \sqrt{ \sum_{k=1}^{K=3} \left( v_k^{(s)}(\textbf{x}_j) - v_k^{(s)}(\textbf{x}_q) \right) ^2}.
    \label{eq:dist}
\end{equation}



\subsection{Counterfactual Explanations for Response Curves} \label{sec:CFE}

Given a sample $\textbf{x}_j$, let $\textbf{x}_j' = \{ x'_{j1}, \dots, x'_{jn}\}$ denote a counterfactual explanation and let $\tilde{R}_s(\textbf{x}'_j)$ be its corresponding aligned approximated response curve.
More specifically, we define $\textbf{x}_j'$ as a data point whose response curve $\tilde{R}_s(\textbf{x}'_j)$ has a different shape from that of $\tilde{R}_s(\textbf{x}_j)$ (i.e., it shows different responsivity), such that $d_s(\textbf{x}_j, \textbf{x}_j') \geq \epsilon$, where $\epsilon$ is a hyperparameter threshold.
The CFE $\textbf{x}'_j$ is obtained by introducing perturbations to the set of passive features:
\[
  x'_{ji} =
    \begin{cases}
      x_{ji} + \Delta_{ji}, & \text{if $i \neq s$}\\
      x_{ji}, & \text{if $i=s$}.
    \end{cases}
\] 
Moreover, we aim to find the minimum subset of passive features that should be affected by small perturbations so that the CFE's responsivity is sufficiently different from that of the original one.
As such, the counterfactual search problem can be cast as a MOO problem that is solved for each sample $\textbf{x}_j$ as follows:
\begin{equation}
\min_{\textbf{x}'_j} \textbf{g}(\textbf{x}_j) = \min_{\textbf{x}'_j} \left( g_1(\textbf{x}_j, \textbf{x}_j'), g_2(\textbf{x}_j, \textbf{x}_j'), g_3(\textbf{x}_j, \textbf{x}_j') \right),
    \label{eq:opt}
\end{equation}
where $g_1(\textbf{x}_j, \textbf{x}_j')$, $g_2(\textbf{x}_j, \textbf{x}_j')$, and $g_3(\textbf{x}_j, \textbf{x}_j')$ are independent objective functions whose goals may contradict.

The first objective maximizes the distance between the transformed response curves using Eq.~\ref{eq:dist}:
\begin{equation}
    g_1(\textbf{x}_j, \textbf{x}_j') = 
    \begin{cases}
      - d_s(\textbf{x}_j, \textbf{x}_j'), & \text{if $d_s(\textbf{x}_j, \textbf{x}_j') < \epsilon$}\\
      - \epsilon, & \text{if $d_s(\textbf{x}_j, \textbf{x}_j') \geq \epsilon$}.
    \end{cases} 
    \label{eq:g1}
\end{equation}
The second objective function is used to minimize the number of features that are modified, which are calculated using the $L_0$ norm of $(\textbf{x}_j - \textbf{x}'_j)$ so that:
\begin{equation}
    g_2(\textbf{x}_j, \textbf{x}_j') = || \textbf{x}_j - \textbf{x}'_j ||_0 = \sum_{i=1}^n \mathbb{I}_{x_{ji} \neq x'_{ji}}.  
    \label{eq:g2}
\end{equation}
Finally, the third objective function minimizes the distance between $\textbf{x}_j$ and $\textbf{x}'_j$:
\begin{equation}
    g_3(\textbf{x}_j, \textbf{x}_j') = \frac{1}{n} \sum_{i=1}^n \delta_G(\textbf{x}_{ji}, \textbf{x}_{ji}'), 
    \label{eq:g3}
\end{equation}
where $\delta_G$ represents the Gower distance, which takes into account that passive features can be numerical or categorical:
\begin{equation}
    \delta_G(\textbf{x}_{ji}, \textbf{x}_{ji}') = 
    \begin{cases}
      \frac{1}{r_j}(|\textbf{x}_j - \textbf{x}_j'|), & \text{if numerical}\\
      - \epsilon, & \text{if categorical},
    \end{cases}
\end{equation}
and $r_i$ indicates the range of values of the $i$-th feature.
All of the datasets used in this work consist of numerical features only (i.e., integer and real-valued); however, the use of the Gower distance in Eq.~\ref{eq:g3} allows us to consider more general cases.

Note that calculating $g_1(\textbf{x}_j)$, $g_2(\textbf{x}_j)$, and $g_3(\textbf{x}_j)$ involves generating the aligned approximated response curves $\tilde{R}_s(\textbf{x}_j)$ and $\tilde{R}_s(\textbf{x}'_j)$.
Some approaches optimize the CFE objective functions as part of a differentiable loss function~\cite{CFEproto}; 
however, those approaches cannot be applied in our case given that generating an aligned approximated response curve is not a differentiable process (see Eq.~\ref{eq:response_curve} and \ref{eq:align}).

Similar to the work proposed by Dandl \textit{et al.}~\cite{CFEMOO}, we use the Non-dominated Sorting Genetic Algorithm II (NSGA-II)~\cite{nsga2} to solve the mixed-variable optimization problem given in Eq.~\ref{eq:opt}. 
NSGA-II is an elitist genetic algorithm that finds Pareto non-dominated solutions to MOO problems and uses a crowding distance measure to maintain diversity in subsequent generations.
NSGA-II was preferred over the follow-on NSGA-III since it has shown superior performance on optimization problems with three objectives~\cite{nsga3}. 
Note that even though we use a similar optimization framework to that of Dandl \textit{et al.}~\cite{CFEMOO}, our objectives are designed to alter the shape of the response curve of a given sample (Fig.~\ref{fig:intro}), while theirs are designed to alter a single response value in the context of classification or regression problems.

Let us consider a population size of $T_0$ CFE candidates, from which NSGA-II may select $T$ non-dominated solutions ($T \leq T_0$) denoted as $\{ {\textbf{x}'}_{j}^{(1)}, \dots, {\textbf{x}'}_{j}^{(T)}\}$.
For each solution, we calculate its performance $z_t = \{ g_1({\textbf{x}'}_{j}^{(t)}), g_2({\textbf{x}'}_{j}^{(t)}), g_3({\textbf{x}'}_{j}^{(t)}) \}$.
The objective space is defined as the set of $T$ three-dimensional points $\textbf{Z} = \{ z_1, \dots, z_T \}$.
A solution $z_{t}$ is said to dominate another solution $z_{q}$ ($z_{t} \preceq z_{q}$) if it is no worse than $z_{q}$ (i.e., $g_1({\textbf{x}'}_{j}^{(t)}) \leq g_1({\textbf{x}'}_{j}^{(q)})$,  $g_2({\textbf{x}'}_{j}^{(t)}) \leq g_2({\textbf{x}'}_{j}^{(q)})$, and $g_3({\textbf{x}'}_{j}^{(t)}) \leq g_3({\textbf{x}'}_{j}^{(q)})$) and it is strictly better than $z_{q}$ in at least one objective. 
By definition, the set of non-dominated solutions $\textbf{Z}$ constitutes a Pareto set on the Pareto front.

The remaining question is how to select the best solution from $\textbf{Z}$.
Recall that we are mainly interested in minimizing the number of modified features that are sufficient to alter the responsivity of $\textbf{x}_j$ such that $d_s(\textbf{x}_j, \textbf{x}_j') \geq \epsilon$.
Thus, from the subset of solutions in the Pareto set that produces the lowest $g_1$ value, we select the solution that yields the lowest $g_2$ value. 
Note that, according to this criterion, we could reconfigure our MOO problem to optimize $g_1$ and $g_2$ only, or $g_1$ and $g_3$ only (selecting the solution with the fewest changes).
Nevertheless, in practice, we noticed substantially faster convergence rates by optimizing the three functions simultaneously, as they were shown to guide the search more effectively.

\subsection{Local and Global Explanations} \label{sec:int}

Two types of explanation are sought: local and global. 
Local explanations convey which passive features have the greatest impact on the responsivity of a given sample, while
global explanations allow for the identification of the passive features with the highest impact on the shape of the response curves generated for a sensitive system in general.

Given an input $\textbf{x}_j$, its local explanation $\alpha_j$ is the set of passive features that were modified during the generation of the counterfactual sample $\textbf{x}'_j$:
\begin{equation}
\alpha_j = \{ i \; | \; (x_{ji} \in \textbf{x}_j)\wedge (x'_{ji} \in \textbf{x}'_j) \wedge (x_{ji} \neq x'_{ji}) \}.
    \label{eq:local}
\end{equation}
On the other hand, the global explanation of a sensitive system is twofold. 
First, we assess individual feature relevance $r_i$ by providing the ratio of times that a feature was modified during the CFE generation process of all samples in $\textbf{X}$:
\begin{equation}
r_i = \frac{1}{N} \sum_{j=1}^N \mathbb{I}_{i \, \in \, \alpha_j}.
    \label{eq:global}
\end{equation}
We acknowledge that passive features are not necessarily independent and thus individual relevance scores are not enough to understand how they interact. 
For that reason, we also report the five most repeated feature combinations.
By doing so, we identify which features react together and which feature combinations are the most effective.
We restrict the number of reported feature combinations to five for conciseness, as explanations becomes more manageable for humans when smaller result sets are provided~\cite{XAIreview}.
However, note that more feature combinations might need to be provided when dealing with problems with several passive features (the datasets used in this work require less than eight).

\section{Experimental Results} \label{sec:results}

For our experiments, we evaluated our approach on a synthetic dataset with 1-D inputs and two real-world crop yield prediction datasets with 2-D outputs.
As stated previously, no other works have been published on the analysis of the impact that a set of passive features have on the shape of response curves generated for an active feature.
Therefore, we were unable to include other methods for comparison.

We also note that the described approach should not be confused with a sensitivity analysis, which is used to study how the different values of a set of independent variables affect the response variable.
In contrast, we study how the different values of a subset of the input features (i.e., the passive features) affect how the response variable reacts to the entire range of admissible values of a selected feature (i.e., the active feature).
Furthermore, we do not assume independence among features; thus, we report the feature combinations with the greatest responsivity impact in addition to the estimated individual feature relevance values.

A 10-fold cross-validation (CV) design was used with all datasets.
Having selected an active feature, the trained network was used to generate the response curves for all samples in the dataset along with their corresponding CFEs.
We argue that including samples from the training set in this process does not lead to biased results.
The reason is that, when generating a response curve, we synthesize samples that were not observed either in the training set or in the validation set.
We considered that the resulting global explanations would be more consistent if they were calculated using as many approximated response curves as possible.
Thus, these curves were used to produce local and global explanations,
and we produced an independent set of explanations for each of the ten folds, given each fold could yield different curves since we have no ground truth for the curves themselves.
Even so, if the data available is sufficiently large and diverse, it is reasonable to expect similar curves and explanation results from the ten models.
The implementation code is available at \url{https://github.com/GiorgioMorales/ResponsivityAnalysis}.

\subsection{Synthetic Dataset} \label{sec:synth}

Validation of the analysis proposed in this work is challenging if the target response curves are unknown, which is the case when working with our real-world agricultural applications.
For this reason, we created a synthetic dataset consisting of 10,000 samples derived from the following multiple non-linear regression problem with five input features:
\begin{equation}
    y = \text{sigmoid}((10 x_1 - 5) + x_2) x_3^2 x_4 + 10 x_5,
    \label{eq:synth}
\end{equation} 
where $x_1 {\sim}\, U(0, 1)$, $x_2 {\sim}\, U(-3, 3)$, $x_3 {\sim}\, U(1, 2)$, $x_4 {\sim}\, U(1, 2)$, and $x_5 {\sim}\, U(0, 2)$.
We used $x_1$ as the active feature ($s=1$).

\begin{figure}[!t]
\centering
    \includegraphics[width=8cm]{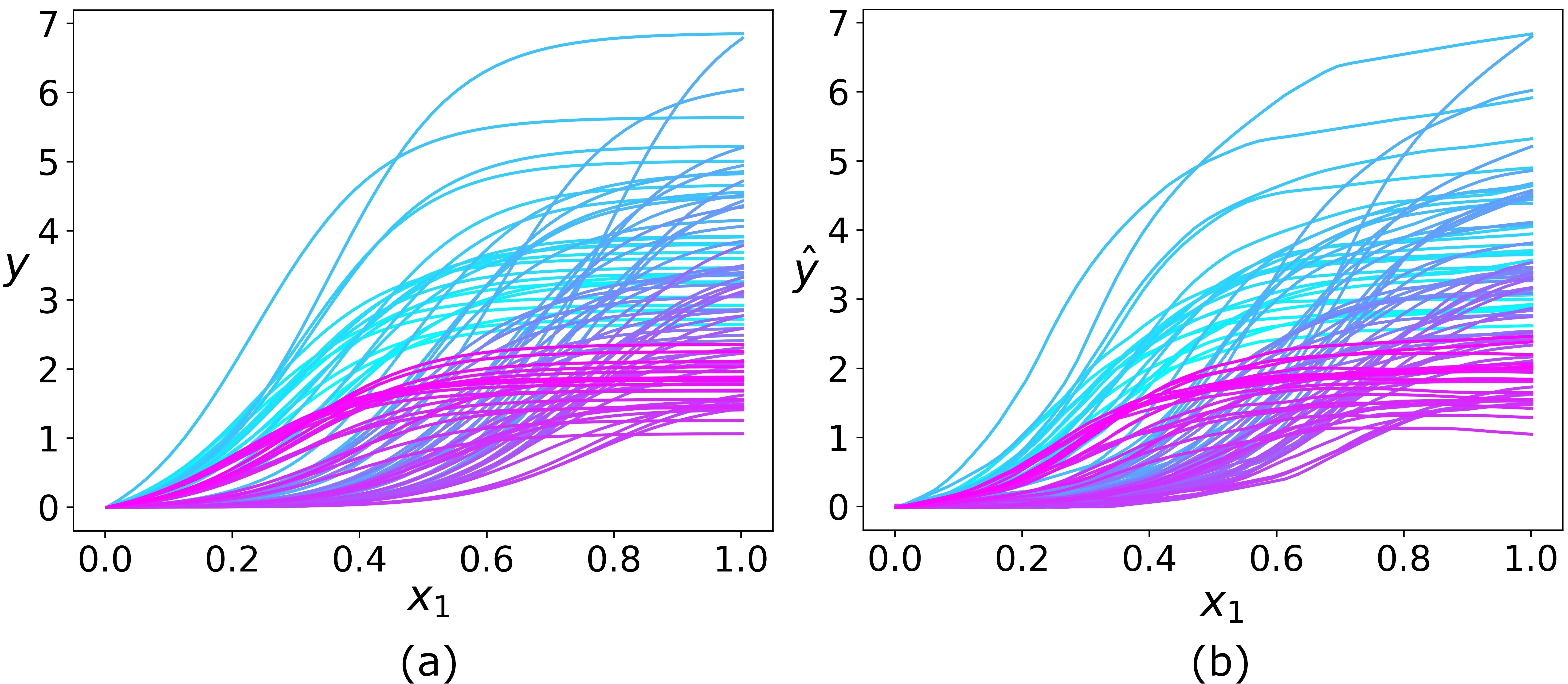}
    \vspace{-1ex}
    \caption{Response curve generation from the synthetic dataset. \textbf{(a)} Ground-truth response curves. \textbf{(b)} NN-generated response curves.}
    \label{fig:synth_curves}
\end{figure}

For these experiments, we trained a feed-forward neural network with two hidden layers, each with 100 nodes.
We calculated the mean square error (MSE) on the validation sets after CV to analyze regression performance.
The resulting average MSE and standard deviation were $6.78 \times 10^{-3} \pm 1.32 \times 10^{-4}$.
Fig.~\ref{fig:synth_curves} shows the aligned ground-truth response curves (generated from Eq.~\ref{eq:synth}) and the aligned approximated response curves using the NN trained during the first CV iteration, for 100 random samples. 

For the CFE generation process, we used a population size of $T=50$ samples for NSGA-II and 100 iterations.
Eq.~\ref{eq:g1} specifies that two response curves show different responsivity if their distance in the transformed space (after using fPCA) is greater than a threshold $\epsilon$.
Since selecting $\epsilon$ is subjective, we considered using multiple threshold values and evaluating the consistency of the results.
Intuitively, higher $\epsilon$ leads to bigger differences between the shapes of two response curves; thus, the number of modified passive features might increase. 
In the future, we plan to replace this threshold with tests that determine if the responsivity of two or more response curves is statistically significant. 
For this dataset, we considered three thresholds: $\epsilon \in \lbrace0.4, 0.6, 0.8\rbrace$.
These values were chosen because they allowed us to see how the feature relevance values change as the $\epsilon$ values increase. 

\begin{figure}[!t]
\centering
    \includegraphics[width=8.5cm]{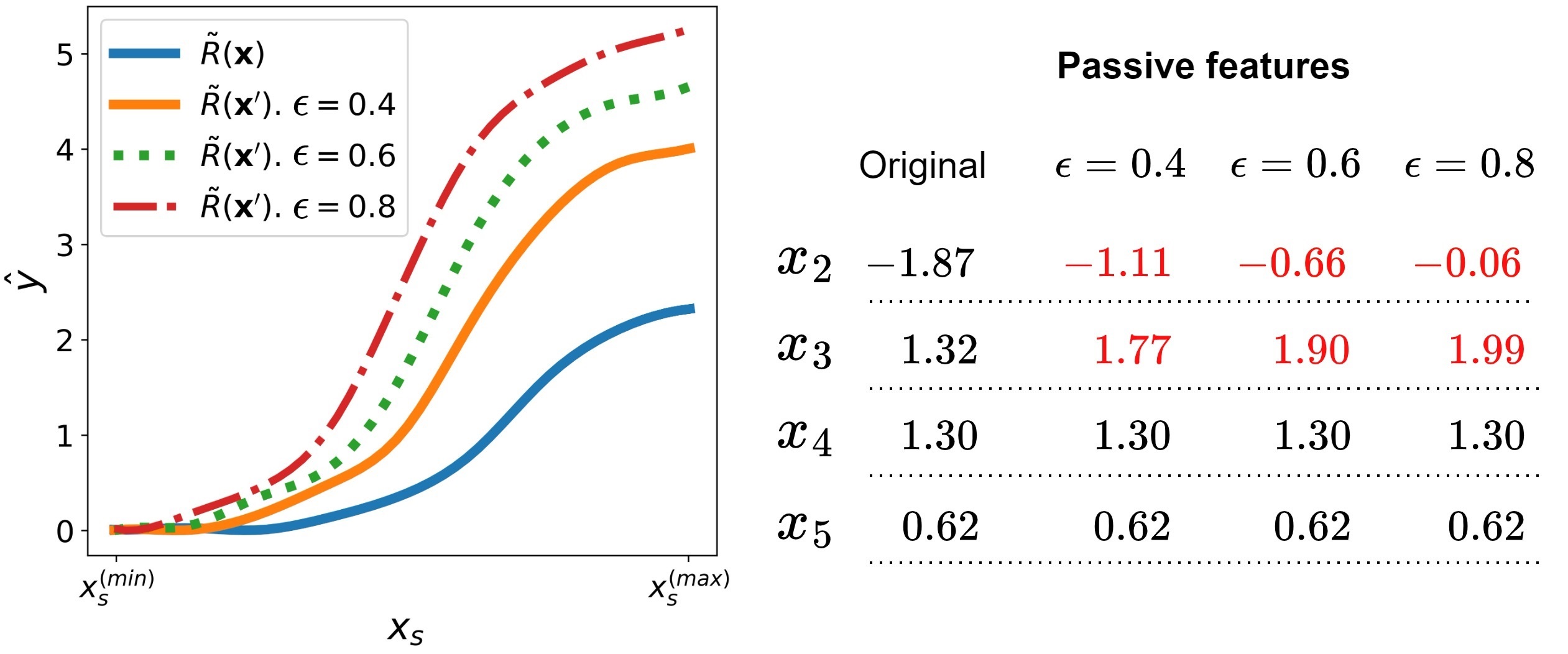}
    \vspace{-1.5ex}
    \caption{Example of the counterfactual response curves generated using $\epsilon = 0.4, 0.6, \text{and}\, 0.8$ for a sample of the synthetic dataset.}
    \label{fig:synth_CFE}
\end{figure}

For example, Fig.~\ref{fig:synth_CFE} shows the counterfactual response curves generated for the first sample of the dataset ($j=1$) using the network trained during the first CV iteration.
Given this input, the local explanation is given by $\alpha_1 = \{ 2, 3 \}$ for the three selected thresholds.
In other words, it was sufficient to modify the second and third features, $x_2$ and $x_3$ to alter the responsivity of the selected sample, regardless of the threshold.

\begin{figure}[!t]
    \centering
    \includegraphics[width=8.5cm]{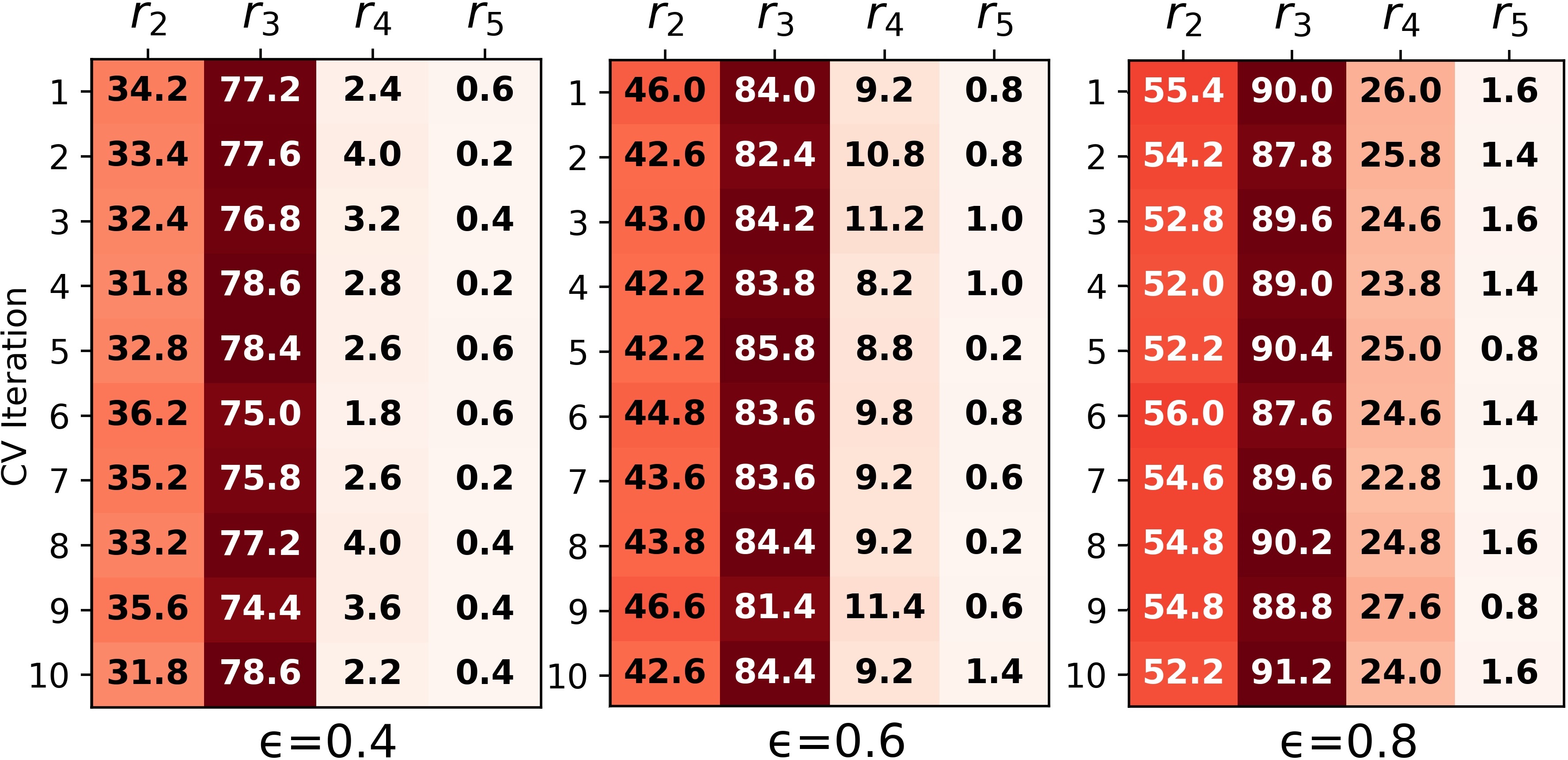}
    \vspace{-1ex}
    \caption{Individual relevance of passive features (in \%) of the synthetic dataset.}
    \label{fig:synth_global}
    \vspace{-1ex}
\end{figure}

We repeated this process for all the samples to obtain global explanations.
Fig.~\ref{fig:synth_global} shows the individual feature relevance values achieved for all CV iterations and $\epsilon$ values.
The remaining global explanations are given by the most frequent feature combinations, which are reported in Table~\ref{tab:synth_comb}.
Specifically, we counted the combination of features that were most repeated across all the CV iterations and selected the top five.
For each selected combination, we calculated the ratio of times it appeared in each CV iteration and reported the mean ratio along with its corresponding standard deviation.

\begin{table}[!t]
    \centering
    \large
    \caption{Top-five feature combinations -- Synthetic dataset}
    \label{tab:synth_comb}
    \vspace{-1ex}
    \resizebox{\columnwidth}{!}{
    \def\arraystretch{1.2}%
    \begin{tabular}{|c|cc|cc|cc|}
    \hline
    \textbf{$\epsilon$} & \multicolumn{2}{c|}{\textbf{$0.4$}} & \multicolumn{2}{c|}{\textbf{$0.6$}} & \multicolumn{2}{c|}{\textbf{$0.8$}} \\ \hline
    \# & \multicolumn{1}{c|}{Comb.} & \% Rep. & \multicolumn{1}{c|}{Comb.} & \% Rep. & \multicolumn{1}{c|}{Comb.} & \% Rep. \\ \hline
    \textbf{1} & \multicolumn{1}{c|}{{[}$x_2, x_3${]}} & $10.9 \pm 0.6$ & \multicolumn{1}{c|}{{[}$x_2, x_3${]}} & $27.2 \pm 1.2$ & \multicolumn{1}{c|}{{[}$x_2, x_3${]}} & $36.6 \pm 1.2$ \\ \hline
    \textbf{2} & \multicolumn{1}{c|}{{[}$x_3, x_4${]}} & $2.7 \pm 0.7$ & \multicolumn{1}{c|}{{[}$x_3, x_4${]}} & $8.3 \pm 0.9$ & \multicolumn{1}{c|}{{[}$x_3, x_4${]}} & $15.5 \pm 1.0$ \\ \hline
    \textbf{3} & \multicolumn{1}{c|}{{[}$x_3, x_5${]}} & $0.2 \pm 0.1$ & \multicolumn{1}{c|}{{[}$x_2, x_3, x_4${]}} & $0.6 \pm 0.4$ & \multicolumn{1}{c|}{{[}$x_2, x_3, x_4${]}} & $6.4 \pm 0.8$ \\ \hline
    \textbf{4} & \multicolumn{1}{c|}{{[}$x_2, x_5${]}} & $0.1 \pm 0.1$ & \multicolumn{1}{c|}{{[}$x_2, x_4${]}} & $0.4 \pm 0.2$ & \multicolumn{1}{c|}{{[}$x_2, x_4${]}} & $2.2 \pm 0.5$ \\ \hline
    \textbf{5} & \multicolumn{1}{c|}{---} & --- & \multicolumn{1}{c|}{{[}$x_3, x_5${]}} & $0.3 \pm 0.2$ & \multicolumn{1}{c|}{{[}$x_3, x_4, x_5${]}} & $0.4 \pm 0.2$ \\ \hline
    \end{tabular}
    }
    \vspace{-1ex}
\end{table}

\subsection{Yield Prediction Dataset} \label{sec:yieldpred}

To consider the usefulness of our approach in a real-world setting, we analyzed data collected on a crop yield prediction problem, which is one of the main tasks of precision agriculture.
Accurate and reliable crop yield prediction provides farmers with tools to make informed decisions, such as determining the nitrogen fertilizer rates needed in specific regions of their fields to maximize their profits~\cite{maxwell}.

We used an early-yield prediction dataset of winter wheat we curated and presented in previous work~\cite{Morales_2023}.
The early-yield prediction is posed as a regression problem where the explanatory variables are represented by a set of features obtained during the growing season (March):
\begin{enumerate}
    \item $N$: Nitrogen rate applied previously (lb/ac).
    \item $A$: Topographic aspect (radians).
    \item $S$: Topographic slope (degrees).
    \item $TPI$: Topographic position index.
    \item $P$: Prior year precipitation (mm).
    \item $VH$ and $VV$: Backscattering coefficients obtained from synthetic aperture radar (SAR) images from Sentinel-I.
\end{enumerate}
The response variable corresponds to the yield value in bushels per acre (bu/ac), measured during the harvest season (August).
Hence, the data acquired in March is used to predict crop yield values in August of the same year.

Each sample $\textbf{x}_j = \{ x_{j1}, \dots x_{jn} \}$ is represented as a spatial data cube of $5 \times 5$ pixels with seven features or channels ($n=7$); that is, $x_{ji} \in \mathbb{R}^{5 \times 5}, \, \forall i \in [1, n]$.
Each pixel represents a region of $10 \times 10\,$\SI{}{\meter} of the field.
The output represents the yield value corresponding to the central pixel of the input sample ($y_j \in \mathbb{R}$).
To tackle this regression problem, we trained a convolutional neural network.
In particular, we use the Hyper3DNetReg 
network architecture we proposed in \cite{Morales_2023}.  
It is a 3D--2D CNN specifically designed to predict the yield values of small spatial neighborhoods of a field.
For our experiments, we used data collected from two winter wheat fields, which we refer to as ``Field A'' and ''Field B''.
Data from three growing seasons were collected for each field (2016, 2018, and 2020).

The selected active feature of this problem was $N$ (i.e., fertilizer input, $s=1$), as we are interested in the analysis of the N-response curves.
The remaining six features constituted the set of passive features.
Important factors such as EONR depend directly on the shape of these curves. 
For instance, EONR is traditionally found as the fertilizer rate at which the first derivative of the N-response curve is equal to a common yield-nitrogen price ratio~\cite{bullock94}.

\begin{figure}[!t]
    \centering
    \includegraphics[width=8cm]{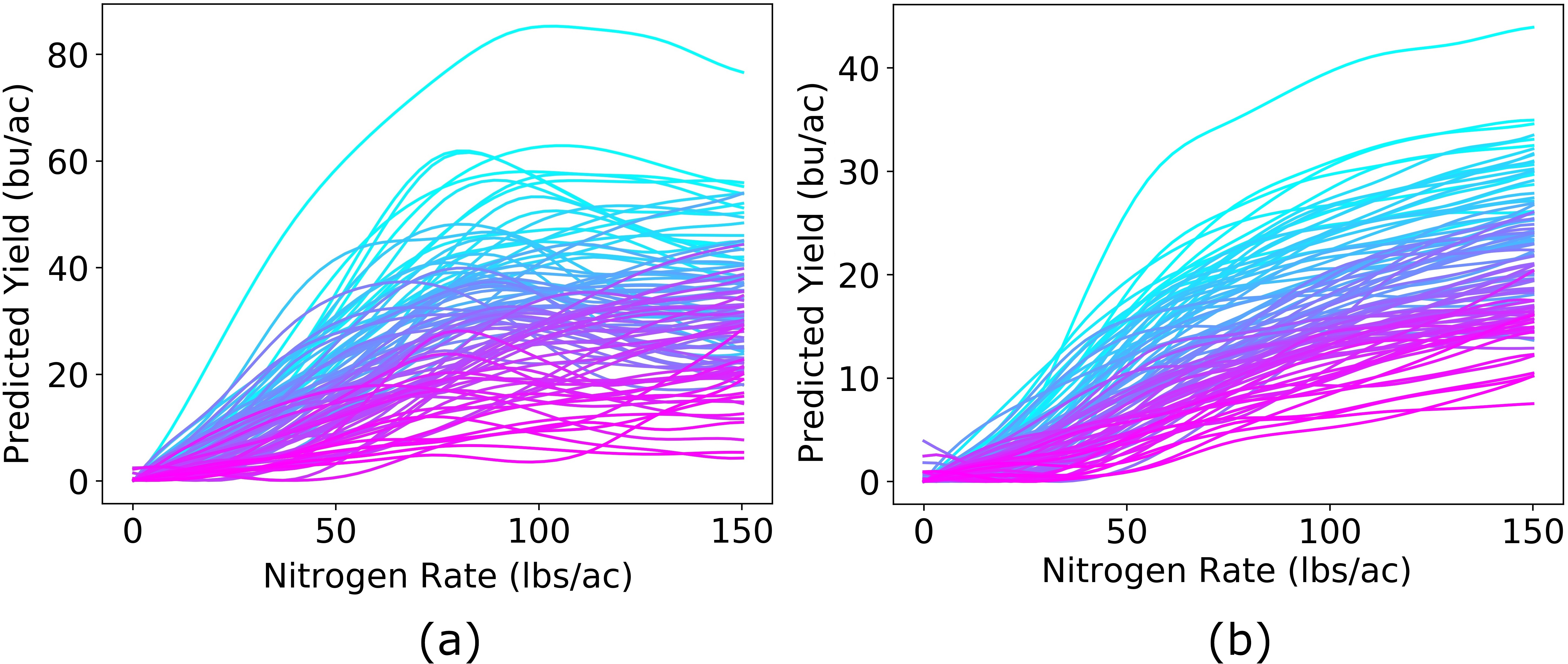}
    \vspace{-1ex}
    \caption{N-response curve generation using 100 random samples from the two yield prediction datasets. \textbf{(a)} Field A. \textbf{(b)} Field B.}
    \label{fig:field_curves}
    \vspace{-1ex}
\end{figure}

\begin{figure}[!t]
    \centering
    \includegraphics[width=8cm]{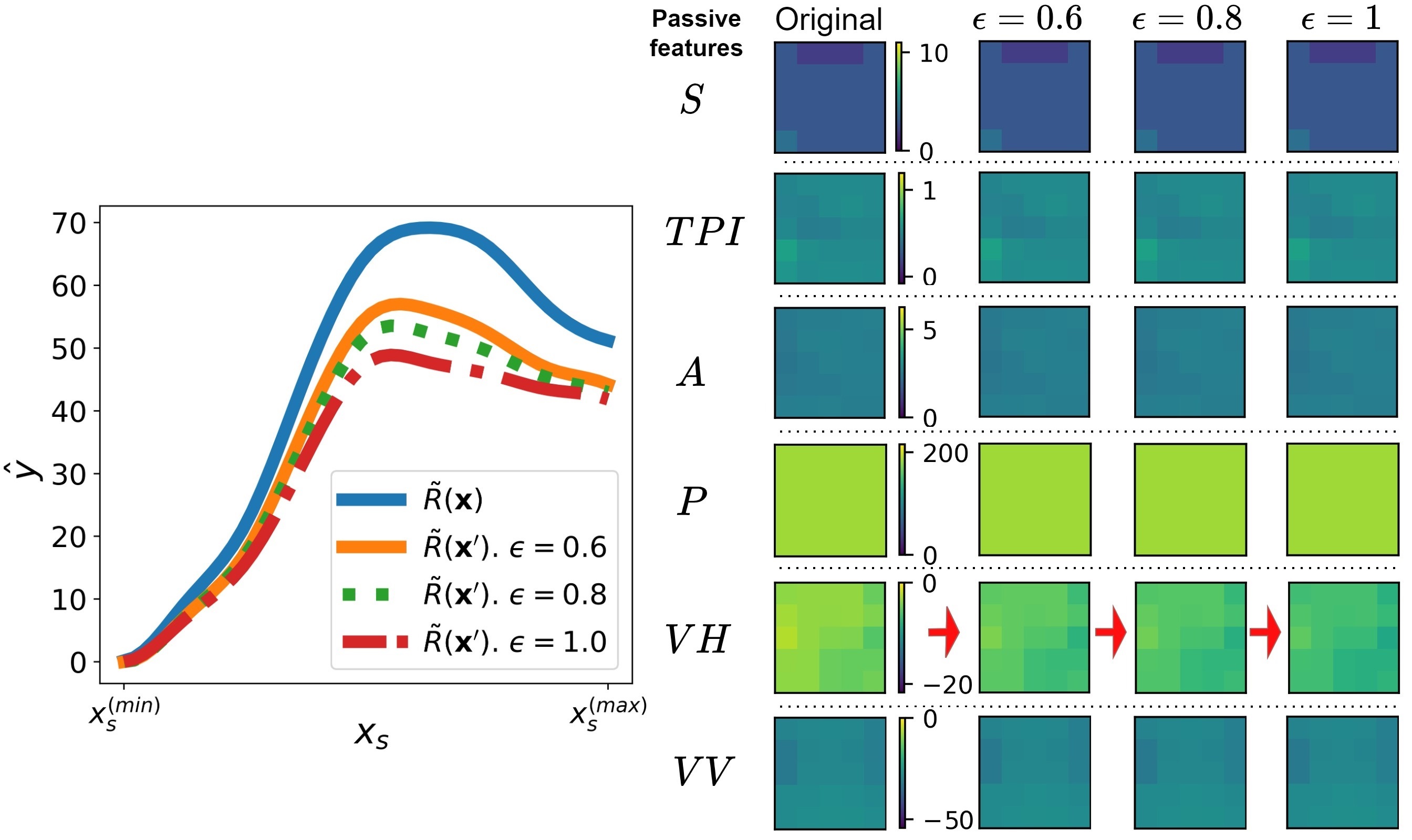}
    \vspace{-1ex}
    \caption{Example of the counterfactual N-response curves generated using $\epsilon = 0.6, 0.8, \text{and}\, 1.0$ for a sample of field A.}
    \label{fig:field_CFE}
\end{figure}

After CV, the average validation MSE and standard deviation were $147.25 \pm 8.17$ and $50.02 \pm 4.29$ for fields A and B, respectively.
Fig.~\ref{fig:field_curves}.a and Fig.~\ref{fig:field_curves}.b show the aligned approximated response curves generated for 100 random samples of field A and B, respectively, using the  Hyper3DNetReg 
network.
For these datasets, we experimented with $\epsilon \in \lbrace 0.4, 0.6, 0.8 \rbrace$.
Fig.\ref{fig:field_CFE} shows an example of the response curves generated for the first sample of field A ($j=1$) using the network from the first CV iteration.
Here, the local explanation is given by $\alpha_1 = \{ 6 \}$, for the three selected thresholds.
That is, we only needed to decrease the value of the sixth feature, $VH$, to alter the responsivity of the selected sample.
Counterfactual response curves and local explanations generated for samples of field B were similar but were omitted due to space limitations. 

\begin{figure}[!t]
    \centering
    \includegraphics[width=8cm]{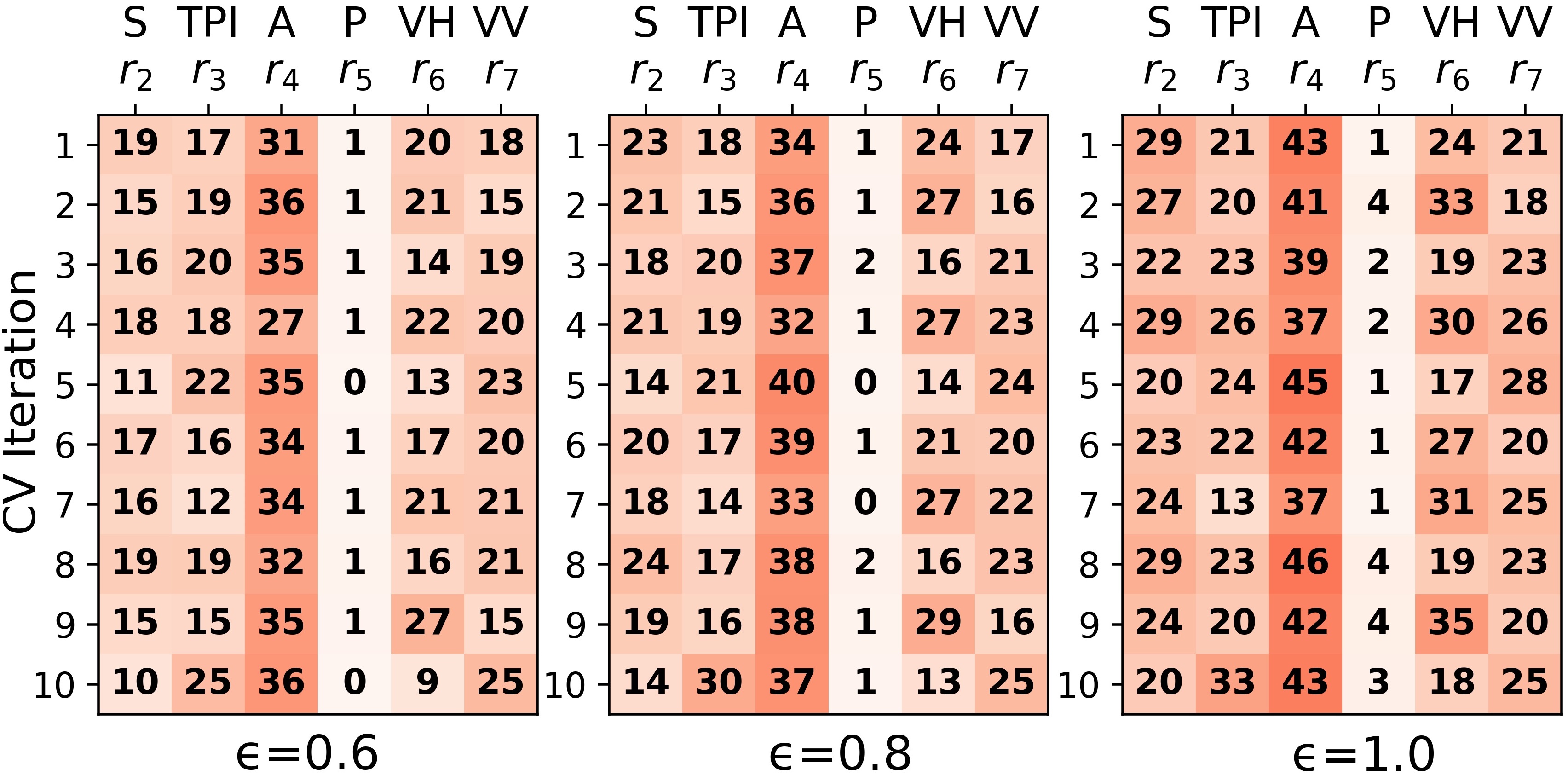}
    \vspace{-1ex}
    \caption{Rounded individual relevance of passive features (in \%) of field A.}
    \vspace{-1ex}
    \label{fig:global_fieldA}
\end{figure}

\begin{figure}[!t]
    \centering
    \includegraphics[width=8cm]{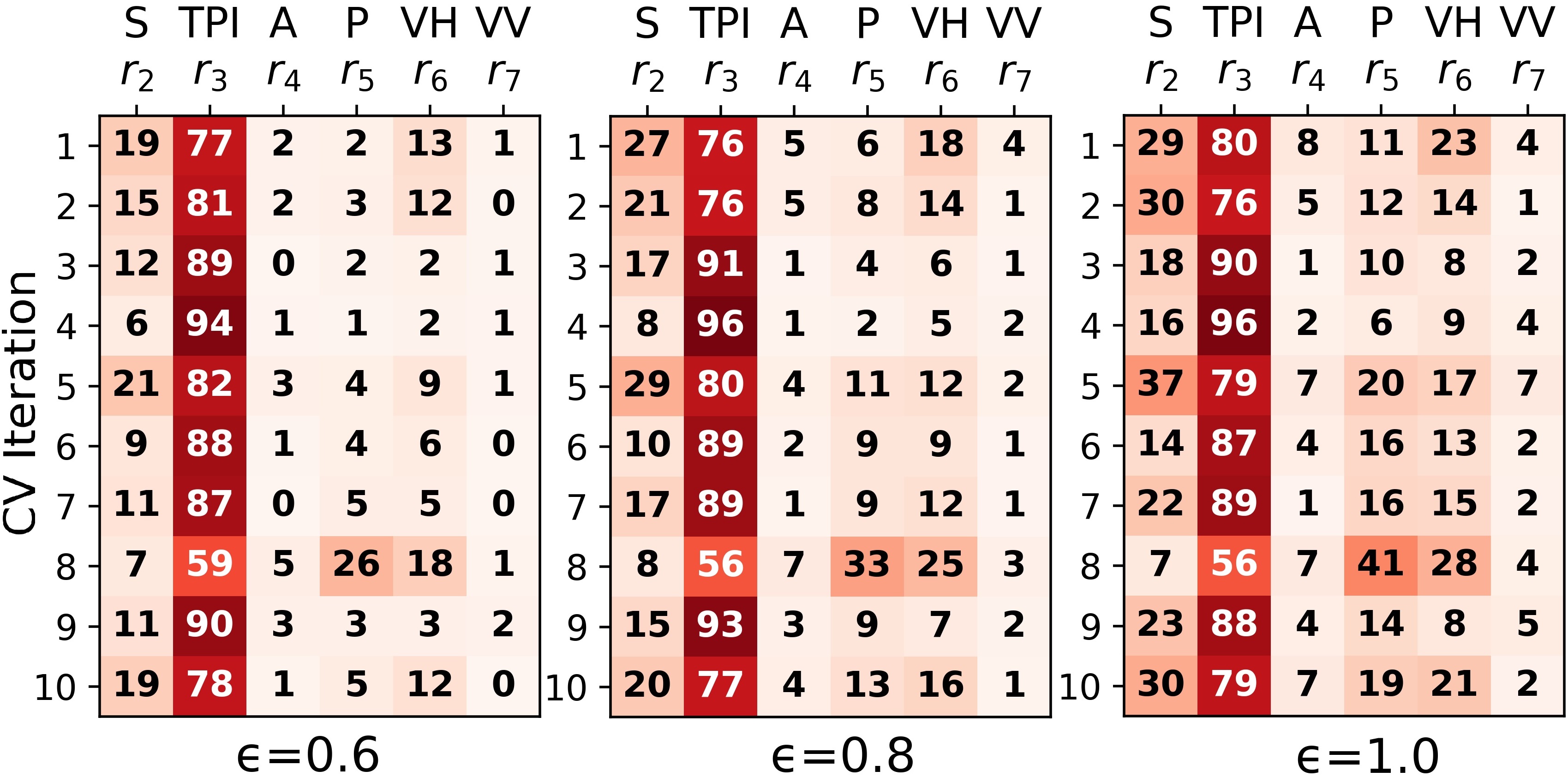}
    \vspace{-1.5ex}
    \caption{Rounded individual relevance of passive features (in \%) of field B.}
    \vspace{-1ex}
    \label{fig:global_fieldB}
\end{figure}

Figs.~\ref{fig:global_fieldA} and \ref{fig:global_fieldB} show the individual feature relevance values achieved for all CV iterations and $\epsilon$ values for fields A and B, respectively.
Tables~\ref{tab:fieldA_comb} and \ref{tab:fieldB_comb} show the mean percent of time that a top-five feature combinations appeared across CV for fields A and B, respectively.

\begin{table}[!t]
    \centering
    \large
    \caption{Top-five feature combinations -- Field A dataset}
    \label{tab:fieldA_comb}
    \vspace{-1ex}
    \resizebox{\columnwidth}{!}{
    \def\arraystretch{1.2}
    \begin{tabular}{|c|cc|cc|cc|}
    \hline
    \textbf{$\epsilon$} & \multicolumn{2}{c|}{\textbf{$0.6$}} & \multicolumn{2}{c|}{\textbf{$0.8$}} & \multicolumn{2}{c|}{\textbf{$1.0$}} \\ \hline
    \# & \multicolumn{1}{c|}{Comb.} & \% Rep. & \multicolumn{1}{c|}{Comb.} & \% Rep. & \multicolumn{1}{c|}{Comb.} & \% Rep. \\ \hline
    \textbf{1} & \multicolumn{1}{c|}{{[}$A, VV${]}} & $0.6 \pm 0.5$ & \multicolumn{1}{c|}{{[}$A, VV${]}} & $1.8 \pm 0.8$ & \multicolumn{1}{c|}{{[}$A, VV${]}} & $4.3 \pm 1.2$ \\ \hline
    \textbf{2} & \multicolumn{1}{c|}{{[}$A, VH${]}} & $0.6 \pm 0.3$ & \multicolumn{1}{c|}{{[}$A, VH${]}} & $1.8 \pm 0.7$ & \multicolumn{1}{c|}{{[}$A, S${]}} & $3.9 \pm 1.1$ \\ \hline
    \textbf{3} & \multicolumn{1}{c|}{{[}$TPI, VV${]}} & $0.5 \pm 0.3$ & \multicolumn{1}{c|}{{[}$S, VH${]}} & $1.6 \pm 0.5$ & \multicolumn{1}{c|}{{[}$A, VH${]}} & $3.3 \pm 0.4$ \\ \hline
    \textbf{4} & \multicolumn{1}{c|}{{[}$TPI, VH${]}} & $0.5 \pm 0.3$ & \multicolumn{1}{c|}{{[}$A, TPI${]}} & $1.6 \pm 0.9$ & \multicolumn{1}{c|}{{[}$S, VH${]}} & $3.2 \pm 0.7$ \\ \hline
    \textbf{5} & \multicolumn{1}{c|}{{[}$S, TPI${]}} & $0.5 \pm 0.3$ & \multicolumn{1}{c|}{{[}$TPI, VH${]}} & $1.5 \pm 0.7$ & \multicolumn{1}{c|}{{[}$A, TPI${]}} & $3.0 \pm 1.4$ \\ \hline
    \end{tabular}
    }
    \vspace{-1ex}
\end{table}

\begin{table}[!t]
    \centering
    \large
    \caption{Top-five feature combinations -- Field B dataset}
    \label{tab:fieldB_comb}
    \vspace{-1ex}
    \resizebox{\columnwidth}{!}{
    \def\arraystretch{1.2}
    \begin{tabular}{|c|cc|cc|cc|}
    \hline
    \textbf{$\epsilon$} & \multicolumn{2}{c|}{\textbf{$0.6$}} & \multicolumn{2}{c|}{\textbf{$0.8$}} & \multicolumn{2}{c|}{\textbf{$1.0$}} \\ \hline
    \# & \multicolumn{1}{c|}{Comb.} & \% Rep. & \multicolumn{1}{c|}{Comb.} & \% Rep. & \multicolumn{1}{c|}{Comb.} & \% Rep. \\ \hline
    \textbf{1} & \multicolumn{1}{c|}{{[}$TPI, S${]}} & $4.4 \pm 1.8$ & \multicolumn{1}{c|}{{[}$TPI, S${]}} & $7.2 \pm 2.7$ & \multicolumn{1}{c|}{{[}$TPI, S${]}} & $9.7 \pm 3.7$ \\ \hline
    \textbf{2} & \multicolumn{1}{c|}{{[}$TPI, VV${]}} & $2.1 \pm 1.2$ & \multicolumn{1}{c|}{{[}$TPI, VV${]}} & $5.0 \pm 1.3$ & \multicolumn{1}{c|}{{[}$TPI, VV${]}} & $6.5 \pm 2.2$ \\ \hline
    \textbf{3} & \multicolumn{1}{c|}{{[}$TPI, P${]}} & $1.9 \pm 1.5$ & \multicolumn{1}{c|}{{[}$TPI, P${]}} & $3.9 \pm 1.7$ & \multicolumn{1}{c|}{{[}$TPI, P${]}} & $5.5 \pm 2.0$ \\ \hline
    \textbf{4} & \multicolumn{1}{c|}{{[}$TPI, A${]}} & $1.0 \pm 0.8$ & \multicolumn{1}{c|}{{[}$TPI, A${]}} & $1.4 \pm 0.5$ & \multicolumn{1}{c|}{{[}$S, P${]}} & $2.3 \pm 1.2$ \\ \hline
    \textbf{5} & \multicolumn{1}{c|}{{[}$S, VV${]}} & $0.6 \pm 0.5$ & \multicolumn{1}{c|}{{[}$S, P${]}} & $1.2 \pm 0.8$ & \multicolumn{1}{c|}{{[}$TPI, A${]}} & $1.7 \pm 0.7$ \\ \hline
    \end{tabular}
    }
    \vspace{-1ex}
\end{table}

\section{Discussion} \label{discusssion}

Although often unobservable, response curves of a system can be approximated using NNs.
Fig.~\ref{fig:synth_curves} suggests that the NN trained on the synthetic dataset learned to generate approximate response curves that were similar in shape to the ground-truth response curves.
Even though ground-truth N-response curves are unobservable, we argue that the curves shown in Fig.~\ref{fig:field_curves} are sound.
For example, previous works have described N-response curves as sigmoid-like curves~\cite{watkins}, similar to most of the curves of Fig.~\ref{fig:field_curves}.b.
Other works have also considered quadratic functions~\cite{bullock94} that account for the apparent decrease in yield response after reaching a certain saturation point, which can be seen in most of the curves of Fig.~\ref{fig:field_curves}.a.
Note that our analysis, as any other explainability method, allows for the explanation of the function approximated by the model.
Specifically, we aim to identify the features (and feature combinations) that the model considers having the most responsivity impact.

The regression problem introduced in Eq.~\ref{eq:synth} consists of four passive features.
$x_2$ alters the shape of the response curves by stretching them horizontally.
$x_3$ multiplies the sigmoid function, as well as $x_4$; however, since it is squared, it is expected to have more impact on the shape of the response curves.
In addition, a small change in $x_3$ produces a vertical stretching so that the resulting distance between the modified and original response curve is greater than the one produced by $x_2$ when modified by the same amount.
From Fig.~\ref{fig:synth_global}, we verified that $x_3$ has greater responsivity impact than $x_4$ and $x_2$, as it was more often modified by our CFE generation process.
Finally, $x_5$ is independent of the rest of the features, which implies that it only shifts the response curves vertically but does not alter their shape.
Again from Fig.~\ref{fig:synth_global}, we verified that $x_5$ was assigned relevance values near $0\%$.
Hence, our method found that the feature with the greatest responsivity impact is $x_3$,
followed in relevance by $x_2$ and $x_4$. 
We conclude that the results obtained by our method coincide with those obtained by the analysis the equation used to generate the dataset.
Also, Table~\ref{tab:synth_comb} shows that the most effective feature combination is $x_2$ and $x_3$, followed by $x_3$ and $x_4$.

It is important to point out that we observed that the relevance ranking of features remained constant for different values of $\epsilon$.
For instance, Fig.~\ref{fig:synth_global} shows the ranking of features (from the most relevant to the least relevant) is: $[x_3, x_2, x_4, x_5]$ for the three tested $\epsilon$ values.
Similarly, Fig.~\ref{fig:global_fieldA} shows that, for the three tested $\epsilon$ values in field A, $A$ is the most relevant feature while the relevance scores of $S$, $TPI$, $VH$, and $VV$ are comparable, and $P$ is the least relevant feature.
A similar behavior is observed in Fig.~\ref{fig:global_fieldB}.
This is meaningful because it suggests that the selection of $\epsilon$ is not crucial when finding the passive features with greater global relevance.

Note in Fig.~\ref{fig:synth_global} that the resulting relevance scores are similar for all CV iterations.
This indicates that the NN model learns similar functions across different iterations.
This is also the case for Fig.~\ref{fig:global_fieldA} and \ref{fig:global_fieldB}, although they show greater variation due to overfitting issues caused by limited data set sizes and lack of data variability in some of the training folds. 


Furthermore, we found that the relevance values increase as $\epsilon$ increases.
This is because, given a sample, if there is an increment in the desired distance threshold, it is likely that the number of modified passive features will grow.
For example, Table~\ref{tab:synth_comb} shows that the most repeated feature combination, $(x_2,x_3)$, occurs 10.9\% of the time when $\epsilon=0.4$ but 36.6\% when $\epsilon=0.8$.
This also happens for fields A and B, as seen in Tables~\ref{tab:fieldA_comb} and \ref{tab:fieldB_comb}.
However, the most repeated feature combination for field A, $(A,VV)$, occurs only 0.6\% of the time when $\epsilon=0.6$ and 4.3\% when $\epsilon=1$.
This means that changes in individual features have greater impact on the models trained for field A.
For instance, in Fig.~\ref{fig:field_CFE}, it was enough to reduce the $VH$ value to alter the shape of the response curve.
This is because $VH$ is related to the moisture content, meaning by lower $VH$ values indicate less moisture in the soil. Thus it is reasonable to expect the soil to be less responsive to higher amounts of nitrogen. 

From Figs.~\ref{fig:global_fieldA} and \ref{fig:global_fieldB}, we see that feature relevance values are different for both fields.
The main reason is that field A is located on steep abrupt terrain while field B is not.
As a consequence, the model trained for field A learned that the aspect $A$ (i.e., the slope orientation) is the most relevant feature.
Interestingly, in terrain with varying elevations located in the Northern Hemisphere, regions that are facing north and east have limited sunlight during the day and are more prone to snow retention.
These are factors that may affect the responsiveness of the fertilizer.
On the other hand, field B has an almost constant elevation, so $A$ is not an important factor. 
The model trained for this field learned that the $TPI$, which is related to the ruggedness of the terrain, is the most relevant feature.
The slope $S$, which influences fertilizer runoff, which in turn affects the responsiveness of the fertilizer, is the second most important factor for this field.
Finally, $P$ was assigned a low responsivity impact on both fields.
This seems counterintuitive considering that precipitation is a critical factor for crop production.
This would suggest that $P$ is independent of the other features and, as in $x_5$ from the synthetic problem, only shifts the response curves vertically but does not affect their shape.


\section{Conclusion} \label{conclusion}

The analysis of feature response curves often ignores other explanatory variables as a source of variability in the shape of the curves.
Acknowledging all relevant variables, quantifying their responsivity impact, and understanding how they interact, may improve the accuracy of important applications such as drug dose optimization and N-fertilizer optimization. 

We presented a method that generates approximate response curves for a selected active feature using neural networks. 
Our approach then estimates the impact that a set of passive features have on the shape of the response curves by generating counterfactual explanations.
Experimental results on a synthetic dataset coincide with expectations following the analysis of the equation that was used to generate the training data.
Experiments on two crop yield prediction datasets found that the factor with the greatest responsivity impact on N-response curves was the terrain aspect, for one of the studied winter wheat fields, and the topographic position index, for the other field.
While perhaps already understood by farmers and agronomists, this analysis confirms that the models are reflecting what science would expect in crop production.
Future work will focus on designing tests that will determine if the responsivity of two or more response curves is statistically significant. 
We also plan to use equation discovery approaches to approximate parametric equations for crop production.

\section*{Acknowledgements}
The authors wish to thank the team members of the On-Field Precision Experiment (OFPE) project for their comments throughout the development of this work.
We would also like to thank Jordan Schupbach for providing advice on the experimental design. 
This research was supported by a USDA-NIFA-AFRI Food Security Program Coordinated Agricultural Project 
(Accession Number 2016-68004-24769), and also by the USDA-NRCS Conservation Innovation Grant from the On-farm Trials Program
(Award Number NR213A7500013G021).

\balance
\bibliographystyle{IEEEtran}
\bibliography{references}

\end{document}